\documentclass[runningheads]{llncs}
\usepackage[T1]{fontenc}
\usepackage{amsmath}
\usepackage{algorithm}
\usepackage{algpseudocode}
\usepackage{graphicx}
\usepackage{subcaption}
\usepackage{lipsum}
\usepackage{multicol}
\usepackage{cite}
\graphicspath{{images/}}
\usepackage{hyperref}
\usepackage{xcolor}
\begin{document}
%
\title{Deep Learning Hyperparameter Optimization for Breast Mass Detection in Mammograms}
%


\author{Adarsh Sehgal\inst{1,2} \and
Muskan Sehgal\inst{3} \and
Hung Manh La\inst{1,2} \and
George Bebis\inst{2}}

%
\institute{Advanced Robotics and Automation (ARA) Laboratory \email{hla@unr.edu} \and
Department of Computer Science and Engineering, University of Nevada, Reno, 89557, NV, USA \and
Reno, Nevada, USA 
}

\maketitle              
\begin{abstract}
Accurate breast cancer diagnosis through mammography has the potential to save millions of lives around the world. Deep learning (DL) methods have shown to be very effective for mass detection in mammograms. Additional improvements of current DL models will further improve the effectiveness of these methods. A critical issue in this context is how to pick the right hyperparameters for DL models. In this paper, we present GA-E2E, a new approach for tuning the hyperparameters of DL models for brest cancer detection using Genetic Algorithms (GAs). 
Our findings reveal that differences in parameter values can considerably alter the area under the curve (AUC), which is used to determine a classifier's performance.

\end{abstract}
\section{Introduction}

The fast growth of machine learning, particularly deep learning, continues to pique the interest of the medical imaging community in using these approaches to increase cancer screening accuracy. Breast cancer is the second-highest cause of cancer death in women in the United States \cite{smith2003american}, and mammography screening has been shown to lower mortality \cite{oeffinger2015breast}. Despite its advantages, screening mammography is linked to a high rate of false positives and false negatives. In the United States, the average sensitivity of digital screening mammography is $86.9\%$, while the average specificity is $88.9\%$ \cite{lehman2017national}. Since the 1990s, computer-assisted detection and diagnosis (CAD) software \cite{elter2009cadx} has been created to assist radiologists in improving the predicted accuracy of screening mammography. Unfortunately, evidence showed that early commercial CAD systems did not result in significant performance improvements \cite{fenton2007influence, cole2014impact, lehman2015diagnostic}, and that progress remained stagnant for more than a decade after their introduction. With deep learning's remarkable success in visual object recognition and detection, as well as many other domains \cite{lecun2015deep}, there is a lot of interest in developing deep learning tools to help radiologists and improve screening mammography accuracy. According to recent studies \cite{rodriguez2019stand, rodriguez2019detection}, a deep learning-based CAD system performed as well as radiologists in solo mode, and even increased radiologists' performance in assist mode.

It has been found that when deep learning is used for detecting breast cancer on screening mammography, it can improve the accuracy to as high as $96\%$. One of the interesting works \cite{shen2019deep}, starts with performing patch classification on a mammogram to find the region of interest (ROI). This patch classifier is then converted to a whole image classifier by introducing a combination of the VGG network \cite{simonyan2014very} and the residual network (Resnet) \cite{he2016deep}. \cite{shen2019deep} uses CBIS-DDSM \cite{lee2017curated} dataset to perform various experiments to generate an end-to-end training process, which we will refer to as E2E. Depending on the combination of top layers used, the authors have generated various models, which can be transferred to other datasets with no ROI information. Since the learning is being transferred, it needs additional tuning of parameters, as mentioned in \cite{shen2019deep}. Even though some of the parameters can be tuned manually, it can be a tedious and inefficient task to train the parameters which have a large number of possible values. With each of the parameters assuming a large  number of possible values, combinations of these values can grow exponentially. This makes it almost impossible to find near-optimal values for the parameters.

Even though the performance of the whole image classifier to detect a mass in a mammogram has been improved, the authors \cite{shen2019deep} do not justify the use of many constant parameters used in transferring the learning to a dataset with minimal/no ROI information. \cite{sehgal2019deep} and \cite{sehgal2019lidar} have shown that automatic tuning of such parameters can greatly enhance the performance of the overall system. \cite{sehgal2019deep} made use of GA with Deep Reinforcement Learning (GA-DRL), and \cite{sehgal2019lidar} used GA with Lidar-monocular visual odometry (GA-LIMO). Both of these works proved that when GA is used for parameter tuning, it has the potential of producing encouraging results. A neural-genetic algorithm for feature selection system was put up by \cite{verma2007novel} to categorize microcalcification patterns in digital mammograms. The outcomes in \cite{verma2007novel} demonstrated that the proposed algorithm was successful in locating a suitable feature subset, which also resulted in a high classification rate. Some other related works include: \cite{sehgal2019genetic}, \cite{sehgal2022ga}, \cite{jiang2007genetic}, and \cite{sehgal2022automatic}. 

In this work, we propose a novel algorithm (called GA-E2E), which uses a Genetic Algorithm (GA) on the E2E model to tune the parameters for transferring the learning to Inbreast database \cite{moreira2012Inbreast}. We are using the pre-trained model files from E2E. These pre-trained models were generated for whole-mammogram classification. Each run of a fitness function performs E2E training on the Inbreast dataset using the parameter values decided by the GA. The output of the fitness function is Area Under the Curve (AUC). We are taking a GA reference from an open-source implementation \cite{benitez2019jmetalpy}. Our tests revealed that GA-E2E is computationally demanding; hence we propose employing a standalone cluster of two PCs connected by a network switch. We used Spark to create this configuration. Our findings demonstrate that GA-E2E contributed to a higher AUC. GA focuses search on the promising portion of the search space, which saves time, computational power, and manual efforts. Open source code is available at \textcolor{orange}{\href{https://github.com/aralab-unr/GA-mammograms}{https://github.com/aralab-unr/GA-mammograms}}. In the next section, we will describe some closely related works.

\section{Background \& Motivation}

\subsection{End-to-End Pipeline}
\begin{figure*}
\centering
  \begin{subfigure}[b]{\linewidth}
  \includegraphics[width=\textwidth,height=5cm]{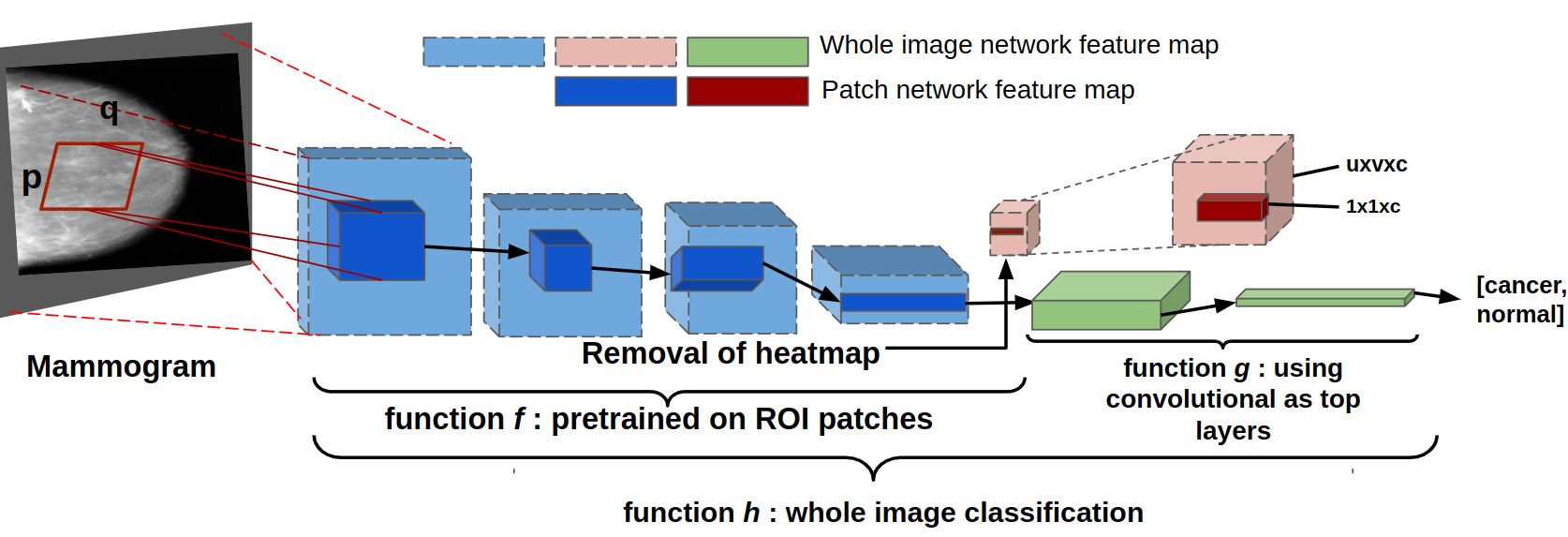}
  \end{subfigure}
  \caption{E2E pipeline.}
  \label{fig:e2epipeline}
\end{figure*}

\subsubsection{Converting a classifier from recognizing patches to whole images}
A common technique for classifying or segmenting large complicated images is to utilize a classifier in sliding window mode to recognize local patches of an image and provide a grid of probabilistic outputs. This is followed by a process that summarizes the outputs of the patch classifier to get the final classification or segmentation result. Using entire slide images of sentinel lymph node biopsies, such approaches have been utilized to detect metastatic breast cancer \cite{wang2016deep} and to segment neuronal membranes in microscopic images \cite{ciresan2012deep}. This technique, however, necessitates two phases, each of which must be tuned separately. Here, training on entire images is performed by combining the two phases into a single step (Fig. \ref{fig:e2epipeline}).

The $h$ function takes complete photos as input and outputs labels for the entire image. As a result, it may be trained from the beginning to the end, giving it two advantages over the two-step technique. First, the entire network may be trained simultaneously, eliminating sub-optimal solutions at each stage; second, the learned network can be transferred to another dataset without relying on ROI annotations explicitly. Large mammography databases with ROI annotations are difficult to come by and very expensive. DDSM, the world's largest public library of ROI annotations for digitized film mammography, contains thousands of pictures with pixel-level annotations that can be used to train a patch classifier. Once the patch classifier has been converted into a whole image classifier $h$, it can be fine-tuned using only image-level labels from other databases. This method significantly reduces the need for ROI annotations, and it has a wide range of applications in medical imaging, including breast cancer diagnosis on screening mammograms.

\subsubsection{Network design}
A simple way to make an entire picture classifier from a patch classifier is to fatten the heat map and connect it to the image classification output with fully-connected layers. After the heatmap, a max-pooling layer can be employed to boost the model's translational invariance to the patch classifier's output. Furthermore, a shortcut between the heatmap and the output can be created to make training easy. The heatmap is derived straight from the output of the patch classifier, which employs the softmax activation:

\begin{gather} 
     f(z)_j = \frac{e^{z_j}}{\sum^{c}_{i=1^{e^{z_i}}}} for \; j =  1,..., c.
    \label{third}
\end{gather}

Convolutional layers, which preserve spatial information, are also employed as top layers. On top of the patch classifier layers, two blocks of convolutional layers (VGG or Resnet) can be added, followed by a global average pooling layer, and finally the image's classification output (Fig. \ref{fig:e2epipeline}). As a result, for full picture classification, this approach provides an "all convolutional" network. The heatmap reduces the depth of the feature map between the patch classifier layers and the top layers, as shown in Fig. \ref{fig:e2epipeline}, potentially causing information loss in the entire picture classification. As a result, when the heatmap was completely removed from the overall image classifier, the upper layers were able to fully utilize the patch classifier's capabilities.

\subsubsection{Developing patch and whole image classifiers on CBIS-DDSM}
The pixel-level annotations for the ROIs are stored in the CBIS-DDSM database, along with their pathologically confirmed labels: benign or malignant. Each ROI is also labeled as a  calcification or mass. The majority of mammograms only had one ROI. All mammograms were converted to PNG format and reduced using interpolation to 1152x896, with no picture cropping. The limitation of GPU memory size prompted the downsizing. By sampling picture patches from ROIs and background regions, two patch datasets were constructed. The size of all patches was 224x224, which was large enough to cover the majority of the ROIs annotated. The first dataset (S1) was made up of two sets of patches, one of which was centered on the ROI and the other was a random background patch from the same image.

\subsubsection{Transfer learning for whole image classification on Inbreast}
The dataset was set up and processed. The Inbreast \cite{moreira2012Inbreast} dataset is a public database that contains FFDM images that were acquired more recently. These images exhibit different intensity profiles than digitized film mammograms from the CBIS-DDSM. As a result, Inbreast is a great way to see if a full picture classifier can be used across different mammography platforms. There are 115 patients and 410 mammograms in the Inbreast database, with both CC and MLO views. Each view is considered a separate image. The radiologists' BI-RADS \cite{timmers2012breast} assessment categories are defined as follows in the Inbreast database: 0: insufficient examination; 1: no findings; 2: benign; 3: probably benign; 4: suspicious; 5: highly indicative of malignancy; 6: known biopsy-proven cancer. \cite{shen2019deep} provides a platform to perform to transfer the learning to any dataset. In this work, we use \cite{shen2019deep} on Inbreast, and show that when the parameters are tuned in an optimal way, it can greatly improve the AUC.

\subsection{Genetic Algorithm}
GAs \cite{davis1991handbook, holland1992genetic, goldberg1988genetic} were developed to investigate poorly-understood areas \cite{de1988learning}, where a comprehensive search is impossible and conventional search methods perform poorly. When GAs are used as function optimizers, their goal is to maximize fitness that is related to the optimization goal. Evolutionary computing approaches in general, and GAs in particular, have had a lot of empirical success on a variety of difficult design and optimization problems. They start with a population of randomly started candidate solutions, which are frequently encoded as a string (chromosome). A selection operator reduces the search space to the most plausible spots, but crossover and mutation operators generate new possibilities.

\subsection{Binary Tournament Selection}
A tournament selection (TS) operator is based on a competition among a group of parents. The fitness of a potential solution is calculated for each parent, and the parent with the highest fitness is chosen. The term "binary tournament" refers to a tournament with a field size of two players, which is the most basic kind of tournament selection \cite{back2018evolutionary}. The binary tournament selection (BTS) process begins with the selection of two people at random. The fitness levels of these people are then assessed. The person with the best fitness is then chosen. The tournament selection has the advantage of being able to handle either minimization or maximization problems without requiring any structural changes. Furthermore, the use of a negative value is unrestricted \cite{wahde2008biologically}.

\subsection{Simulated Binary Crossover (SBX)}
The search power of the simulated binary crossover (SBX) \cite{deb1995simulated} is similar to that of the single-point crossover. The distinction between real-coded GAs with SBX and binary-coded GAs with single-point crossover implementations is that the former method eliminates variable coding and creates a child string from a probability distribution based on the location of the parent strings.

\subsection{Polynomial Mutation}
A polynomial mutation operator with a user-defined index parameter ($\eta_m$) was proposed by Deb and Agrawal (1999). They determined from theoretical research that $\eta_m$ causes a perturbation of $O((b - a) / \eta_m)$ in a variable, where $a$ and $b$ are the variable's lower and upper bounds. They also discovered that on most of the issues they tried, a value of $\eta_m \in [20, 100]$ was sufficient. A polynomial probability distribution is utilized in this operator to perturb a solution in the neighborhood of a parent. The mutation operator adjusts the probability distribution to the left and right of a variable value so that no value outside the specified range $[a, b]$ is formed. The mutated solution $p'$ for a certain variable is constructed for a random number $u$ formed within $[0, 1]$ for a specified parent solution $p \in [a, b]$, as follows:

\begin{gather} 
    p' = 
    \begin{cases}
        p+\Bar{\delta}_L (p-x_i^{(L)}), \; \; for \; u \leq 0.5, \\
        p+\Bar{\delta}_R (x_i^{(U)} - p), \; \; for \; u > 0.5.
    \end{cases}
    \label{fifth}
\end{gather}

\section{Approach}
\subsection{Pre-processing Images}
Inbreast dataset contains images in DICOM format. The dataset comes with a mapping document that maps mammogram image file names to other information such as patient ID, BI-RADS assessment categories, etc. It also provides sample MATLAB code to read the images. We used this code along with our code to split the DICOM images into two classes: positive and negative. We, then, converted these images to PNG format to be used with the E2E algorithm. Only a limited number of converted images were used in our experiments. The images were downsized to 1152×896. Training, validation, and test sub-set splits are described in the following sub-section. Two phases are used in the experiments: GA training, and testing. GA training uses training images to generate a new model built by training the last two layers in E2E. The generated model is then used on validation sub-sets to compute AUC. The output of the fitness function used for GA is the average of AUCs from all epochs. The testing phase is used to compare the GA-found parameters with the original values of the parameters. This phase uses train and validation sub-sets to generate a model, which is then used with the test sub-set to compute the final AUC.

\subsection{Building Spark cluster}
\begin{figure}[ht!]
\centering
  \includegraphics[width=\linewidth,height=5cm]{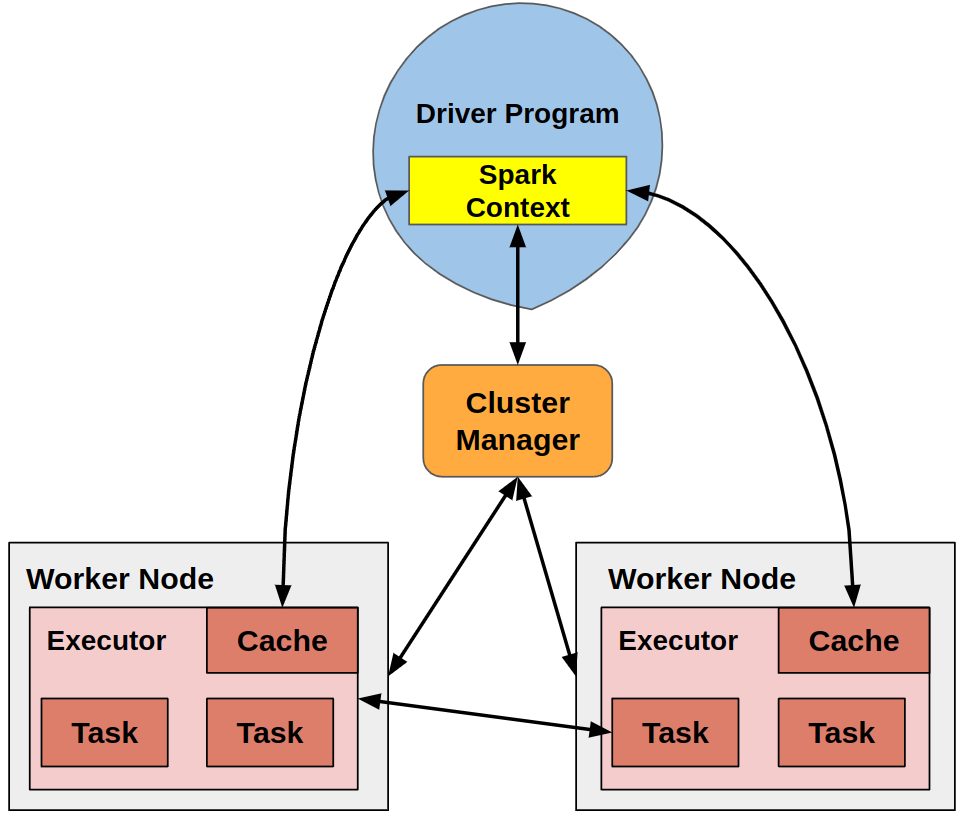}
  \caption{Architecture of Spark cluster.}
  \label{fig:cluster_architecture}
\end{figure}

\begin{figure}[ht!]
\centering
  \includegraphics[width=0.5\linewidth,height=5cm]{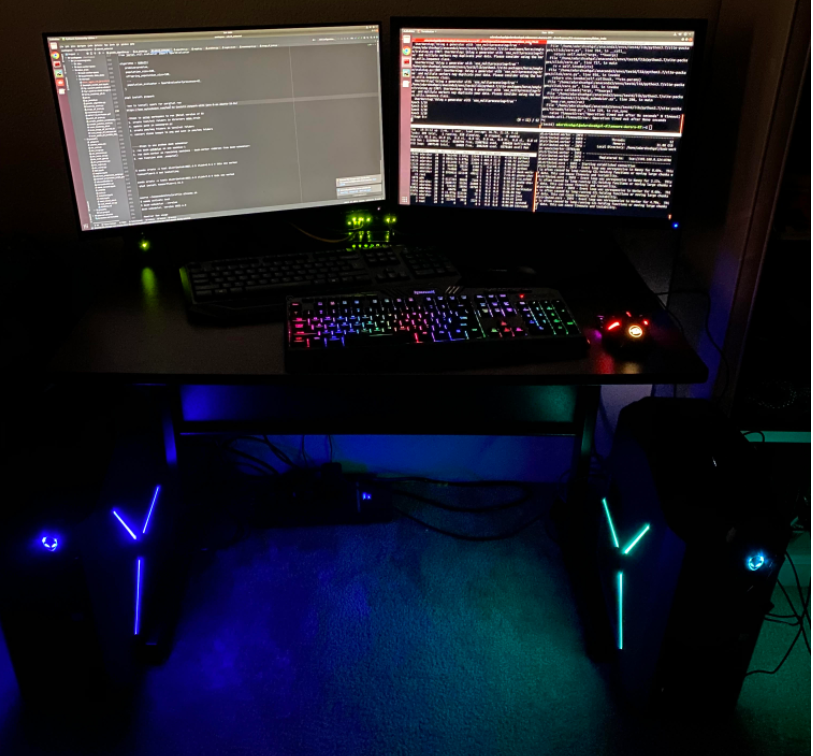}
  \caption{An actual image of computers setup with Spark cluster.}
  \label{fig:cluster_setup}
\end{figure}

\begin{figure}[ht!]
\centering
  \includegraphics[width=0.8\linewidth,height=6cm]{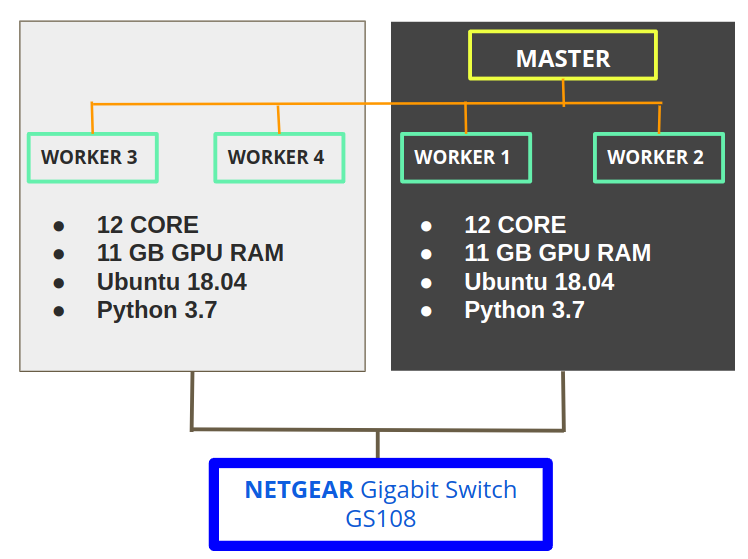}
  \caption{Example representation of master and worker setup for figure \ref{fig:cluster_setup}.}
  \label{fig:cluster_setup_diagram}
\end{figure}

This study uses Spark \cite{zaharia2010spark} to run GA in a distributed setup. This is important because each run of the fitness function is computationally intensive. We used two computers to build a cluster. Advantages of cluster setup include scalability and the  ability for a program to finish in the real time possible. We will describe the details of the GA-E2E algorithm in the next sub-section. The word \textit{cluster} is used to describe the setup of two or more computers running in parallel to achieve a common goal. We chose Spark for our work because it provides easy-to-use configurations, and online resources/documentation. The high-level architecture of Spark is shown in figure \ref{fig:cluster_architecture}. A cluster manager connects \textit{SparkContext} and \textit{Worker Nodes}. For our work, the cluster was built as in figure \ref{fig:cluster_setup}. The details of the corresponding computers are shown in figure \ref{fig:cluster_setup_diagram}. The computers are connected via a network switch. For our implementation specifically, we ran one worker on each of the machines.

\subsection{GA-E2E algorithm}
Now, we set up GA from \cite{benitez2019jmetalpy}. As described in the background section, GA uses BTS, SBX crossover, and polynomial mutation. Since the total training epochs are set at ten, the approximate time for each fitness function evaluation is about 70 seconds for each of the workers. Considering this, we have set the population size to 70, to be run for 70 generations. GA was run on the model from \cite{shen2019deep} using transfer learning on the Inbreast dataset. For each evaluation of the fitness function, GA selects values of parameters, uses them to train the network, computes average validation AUC, and finally uses this trained model on the test sub-set. The AUC returned from the fitness function is the AUC computed on the test sub-set. GA training phases use 30 images in the training sub-set, and 12 images in the validation sub-set. Algorithm \ref{algo:e2ega} describes the high-level overview of the GA-E2E algorithm.

\begin{algorithm}
\caption{Proposed GA-E2E Algorithm}\label{euclid}
\begin{algorithmic}[1]
\State Choose population of $n$ chromosomes
\State Set the values of parameters into the chromosome
\For{all chromosome values} 
    \State Run the code to fine-tune the existing model using independent dataset (Inbreast)
    \State \textbf{return} AUC
\EndFor
\State Perform SBX Crossover
\State Perform Polynomial Mutation
\State Repeat for required number of generations to find optimal solution
\end{algorithmic}
\label{algo:e2ega}
\end{algorithm}

\subsection{Evaluation}

The experiments are broken into two sections, as previously stated: GA training and testing. In both phases, the training and validation sets are identical. However, the final AUC is computed using 100 mammogram images in the testing phase.

\section{Experimental Results}

\subsection{Experimental Setup}
It is important to plan the experiments to be executed to test the algorithm, some details of which we have described in the previous sections. The overall algorithm is described in \ref{algo:e2ega}. The workspace was set up with \cite{shen2019deep}. This code link provides further links to the pre-trained models on whole image classification using the CBIS-DDSM dataset. 

Spark requires that the master and worker be able to communicate key-less using SSH. So, we established password-less SSH between our machines. Both machines can SSH as a user/root into other machines. Both computers have the same username, folder structure, GA-E2E code, same images used in train/validation sub-sets, and the .jar files used by Spark. The number of parallel processes was set to two, one for each worker.

Table \ref{table:e2egaparams} shows the parameters we used in the GA-E2E algorithm. Each set of parameters decided by the GA was used to perform transfer learning on the Inbreast dataset. The reward for the fitness function is the validation AUC averaged over epochs.

\begin{table}[ht]
\centering
\begin{tabular}{|c|c|} 
 \hline
 Parameter & Description\\ 
 \hline
 pos-cls-weight & weight of positive class\\ 
 neg-cls-weight & weight of negative class\\
 weight-decay & stage 1\\
 weight-decay2 & stage 2\\
 init-learningrate & learning rate for stage 1\\
 all-layer-multiplier & multiplier for stages other than stage 1\\
 \hline
\end{tabular}
\caption{Parameters used in the GA-E2E, along with their description.}
\label{table:e2egaparams}
\end{table}

\begin{figure}[ht!]
\centering
  \includegraphics[width=8cm,height=2.2cm]{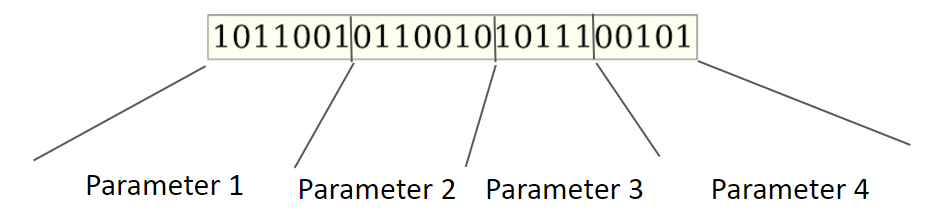}
  \caption{Chromosome representation for the GA.}
  \label{fig:chromosome_rep}
\end{figure}

Each of the parameters is binary encoded into a single chromosome, an example of which is shown in figure \ref{fig:chromosome_rep}. 

From the various manual experiments we did, we would like to mention a few of them. In one such experiment, we used 8 images in a training sub-set, out of which four were negative, and four were positive. By positive we mean that the mammogram image of the breast is malignant, and by negative we mean that the mammogram image of the breast is benign. A similar setup was done for the test sub-set. For the validation sub-set, two images each of the positive, and negative classes were used. Training epochs were fixed to four, and the number of cores of the GPU to be used was fixed to six. The sub-sets chosen had multiple views (CC/MLO) of the same patient. Each epoch took about 160 seconds to complete. On using the learned model on a test sub-set, we got the AUC as 0.5. We experimented with a various number of images in different sub-sets and found that the program ran for a very long time when more images were used. Since GA needs to run the training a few hundred times to find the near-optimal parameter values, for the rest of our implementation, we decided to use the number of images as in the sub-set division described in this section. However, to run GA-E2E, we used only CC views from different patients for training, validation, and testing.

Finally, we used 30 images for training, 12 for validation, and 100 for test sub-sets. Each GA fitness function was run for ten epochs, while the testing phase had 100 epochs. The E2E uses an early stopping approach; hence the training stopped between 30-50 epochs each time the training was run. This was applicable for both original and GA-found parameters. Our experiment assumes each view of the mammogram as a separate image.

\subsection{Running GA-E2E}
The system has now been set up to run the algorithm GA-E2E from \ref{algo:e2ega}. Based on the previously described system setup, the GA completed running in about two days. We captured the logs of each run of the fitness function and the corresponding values for each of the parameters. Table \ref{table:e2egaparamsComp} shows the comparison of values of parameters for E2E and GA-E2E. E2E parameters are the ones originally used in transfer learning with Inbreast in \cite{shen2019deep}. It can be observed that changing parameter values has a significant effect on the AUC of the test sub-set. The increase in AUC is as high as $11\%$ when compared the to the original AUC. 

\begin{table}[ht]
\centering
\begin{tabular}{|p{0.2\linewidth}|p{0.2\linewidth}|p{0.2\linewidth}|p{0.2\linewidth}|} 
 \hline
 Parameter & Range & E2E, AUC=0.51 & GA-E2E AUC=0.57\\ 
 \hline
 pos-cls-weight & 0 - 0.999 & 1.0 & 0.948\\ 
 neg-cls-weight & 0 - 0.999 & 1.0 & 0.319\\
 weight-decay & 0 - 0.999 & 0.0001 & 0.269\\
 weight-decay2 & 0 - 0.999 & 0.0001 & 0.225\\
 init-learningrate & 0 - 0.999 & 0.01 & 0.0008\\
 all-layer-multiplier & 0 - 0.999 & 0.1 & 0.475\\
 \hline
\end{tabular}
\caption{E2E vs. GA-E2E values of parameters.}
\label{table:e2egaparamsComp}
\end{table}

\subsection{Training evaluation}
\begin{figure}[ht!]
\centering
  \includegraphics[width=6cm,height=5cm]{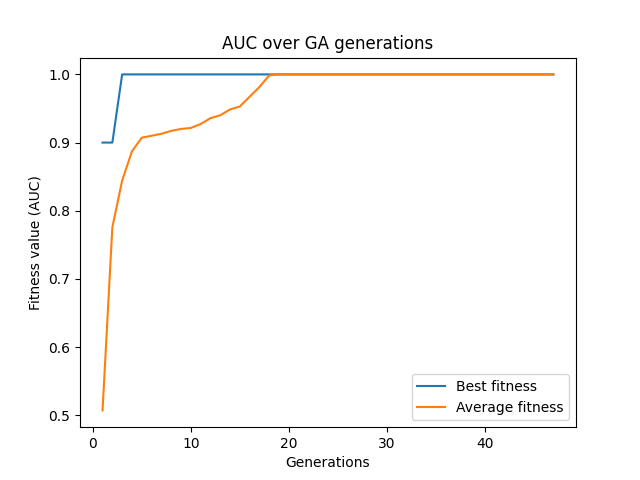}
  \caption{Fitness value evaluation over GA generations.}
  \label{fig:fitness_generations}
\end{figure}

It is significantly important to monitor the progress of a GA as it is running to optimize the performance of the system. From the way GA's work, some of the chromosomes will outperform others. Ultimately, the average fitness would increase towards the maximum fitness value. Figure \ref{fig:fitness_generations} shows the best and average fitness value (AUC) over generations. 

\subsection{Analysis}
It can be observed that all generations except the first two, saw the best fitness value of 1.0. However, the average fitness value started at 0.5, and ultimately, the GA was able to have the entire population at 1.0 AUC. GA took about 18 generations to reach the maximum success rate. Even though the GA reached its maximum success rate, the system was left to run until completion.

\section{Experimental Issues}
Configuring the Spark cluster posed major challenges. Each configuration parameter needs to be appropriately set for the standalone cluster to work using GPUs. The GPU makes program execution much faster, and running the program on a cluster of two computers, cuts the running time to half. Any combination of incorrectly set-up configuration values can result in an unequal load on the machines, which can further lead to some machines being under-utilized/over-utilized.

\section{Conclusion and Future Work}

    \subsection{Conclusion}
    This paper shows that there is no pattern between the values of parameters and the AUC; hence the role of GA is significant in tuning the parameter values. We also show that GA focuses search in the promising search space. GA started with an AUC close to zero. As the GA progressed, the AUC got higher and ultimately surpassed the originally achieved AUC of 0.51. This points to an increase of around $11\%$ in comparison to the original AUC.
    
    \subsection{Future Work}
    We believe that GA may be able to reach the AUC of 1.0 much earlier if GA operators are designed specifically for E2E. Also, different convolution layers can be tested with E2E for better adaptability of the learned model with an independent dataset. Adding more computers to the cluster may reduce execution time further. Modified GA-E2E could be run on a supercomputer, which would enable the ability to perform various experiments using a variety of datasets.

%
%
%
%
\bibliographystyle{splncs04}
\bibliography{root.bib}

\end{document}